\documentclass[runningheads]{llncs}
\usepackage{graphicx}
\usepackage{comment}
\usepackage{amsmath,amssymb} 
\usepackage{color}

\newcommand{\vect}[1]{\boldsymbol{#1}}
\usepackage{algorithm}
\usepackage{algorithmic}
\usepackage{multirow}
\usepackage[table,xcdraw]{xcolor}

\usepackage[colorlinks=true, pdfborder={0 0 0}, linkcolor=green]{hyperref}

\begin{document}
\pagestyle{headings}
\mainmatter
\def\ECCVSubNumber{5462}  

\title{Stochastic Attribute Modeling \\for Face Super-Resolution} 

\titlerunning{Abbreviated paper title}
%
\author{Hanbyel Cho\inst{1} \and
Yekang Lee\inst{1} \and
Jaemyung Yu\inst{1} \and
Junmo Kim\inst{1}}
%
\authorrunning{H. Cho et al.}
%
\institute{School of Electrical Engineering, KAIST, South Korea\\
\email{\{tlrl4658, askhow, jaemyung, junmo.kim\}@kaist.ac.kr}}
\maketitle

\begin{abstract}
When a high-resolution (HR) image is degraded into a low-resolution (LR) image, the image loses some of the existing information. Consequently, multiple HR images can correspond to the LR image. Most of the existing methods do not consider the uncertainty caused by the stochastic attribute, which can only be probabilistically inferred. Therefore, the predicted HR images are often blurry because the network tries to reflect all possibilities in a single output image. To overcome this limitation, this paper proposes a novel face super-resolution (SR) scheme to take into the uncertainty by stochastic modeling. Specifically, the information in LR images is separately encoded into deterministic and stochastic attributes. Furthermore, an Input Conditional Attribute Predictor is proposed and separately trained to predict the partially alive stochastic attributes from only the LR images. Extensive evaluation shows that the proposed method successfully reduce the uncertainty in the learning process and outperforms the existing state-of-the-art approaches.

\end{abstract}

\section{Introduction}
\label{ch:1}
Face super-resolution (SR), also known as face hallucination, is a task in which an input low-resolution (LR) face image is restored to a high-resolution (HR) face image. Face SR is commonly used as a tool to preprocess the degraded input images for other face-related tasks, such as face verification \cite{FaceVerification1,FaceVerification2,FaceVerification3} and face detection \cite{FaceDetection1,FaceDetection2}, as a high performance can be attained when high-quality input images are used.

In recent years, there have been significant achievements in deep neural network (DNN) based SR methods. SRCNN \cite{SRCNN} firstly introduced convolutional neural network (CNN) based SR method, which applied patch extraction and non-linear mapping on the SR processing. VDSR \cite{VDSR} proposed residual architecture so that the network only learns the difference between LR and HR images, and also utilized gradient clipping for fast training. As a special case of the general SR, there exists face related prior knowledge in face images, which can be major clue for face SR problem. To utilize such rich information, GLN \cite{GLN} extracted the global and local spatial information from an LR image, and the HR image was restored using both of information. Wavelet-SRNet \cite{Wavelet-SRNet} did the same thing in frequency domain. Furthermore, to improve the performance of face SR, some reseachers \cite{Super-FAN,FSRNet} utilized various facial priors, which involve semantic meanings such as facial landmarks \cite{FacialLandmarks1,FacialLandmarks2} and parsing maps \cite{ParsingMaps1,ParsingMaps2}.

\begin{figure}[t]
    \centerline{\includegraphics[width=100mm]{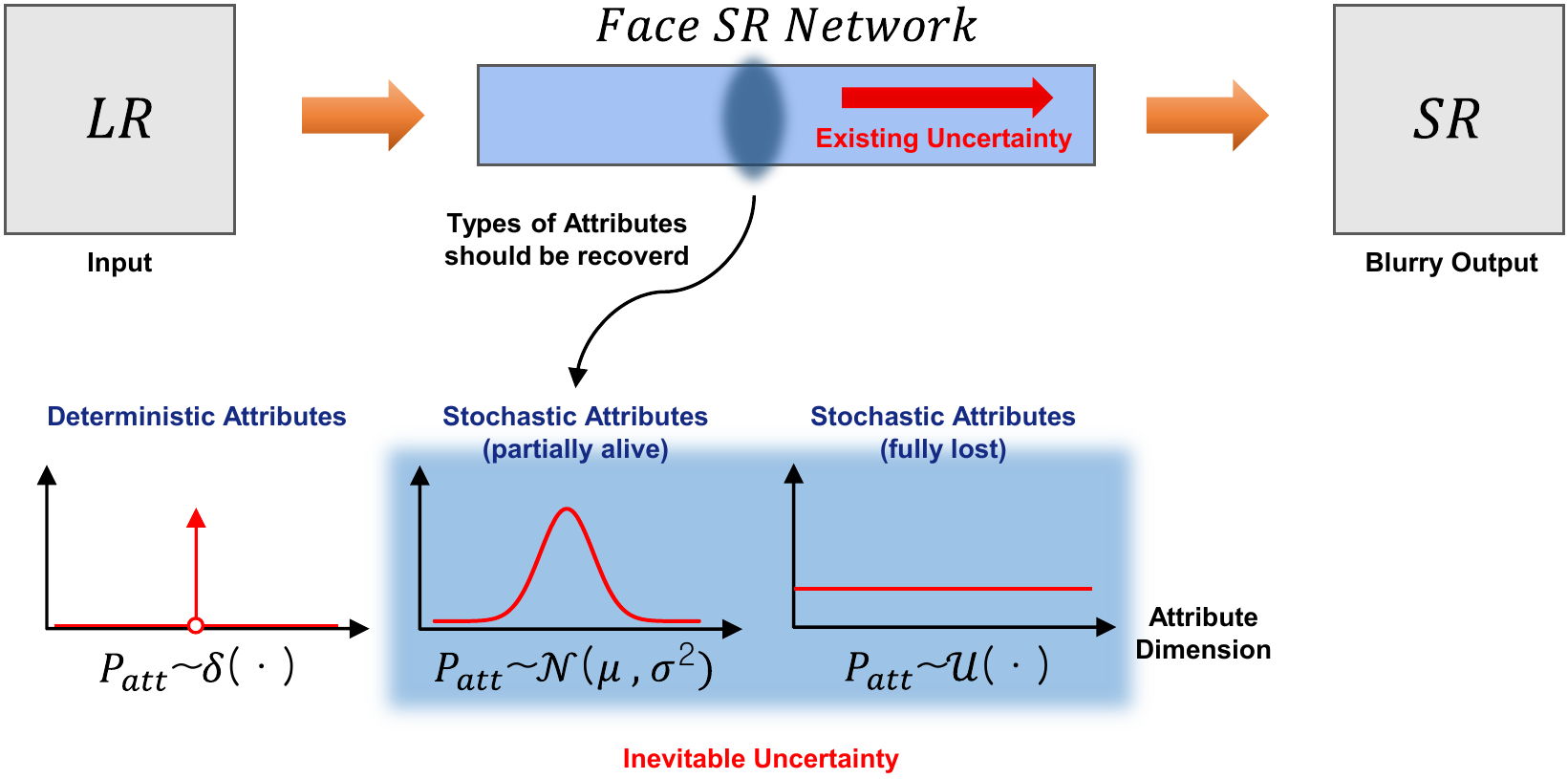}}
    \caption[Information flow in conventional face super-resolution networks]{
    Information flow in conventional face super-resolution networks.
    } \label{fig:Information Flow in FSR}
\end{figure}

\begin{figure}[t]
    \centerline{\includegraphics[width=\textwidth]{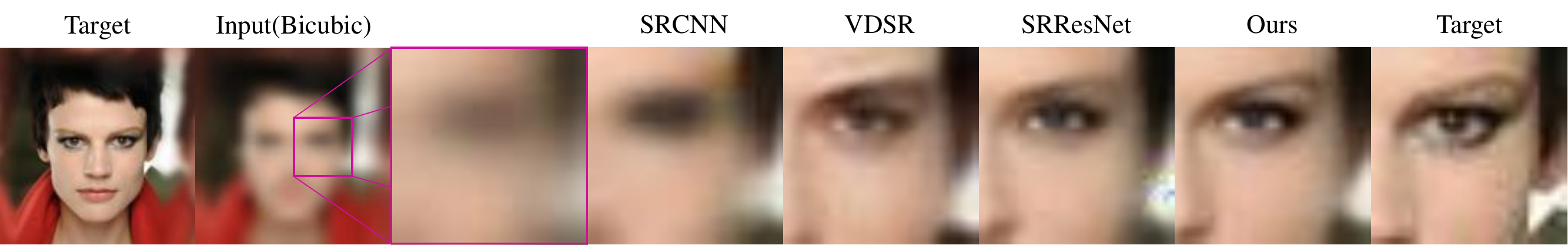}}
    \caption[Blurry outputs of conventional face SR methods for a scale factor of 8]{
    Blurry outputs of conventional face SR methods for a scale factor of 8.
    } 
\label{fig:Blurry Outputs}
\end{figure}

Despite these efforts, the face SR approaches encounter a major challenge of the restored images still being blurry. As shown in Figure \ref{fig:Blurry Outputs}, the existing approaches \cite{SRCNN,VDSR,SRResNet} lead to blurry results in the distinguishing parts of a person such as the eyebrows, because several possible HR images can be inferred from the LR image.

In the face domain, the attributes (such as the shape of the eyebrows) that distinguish the identity of a person can be considered as key features that a network should restore. Such attributes can be classified into two main categories, namely, deterministic and stochastic attributes, depending on the degree of information loss, as shown in Figure \ref{fig:Information Flow in FSR}.

The deterministic attributes are the attributes of an HR image, which can be clearly determined from the LR even if the image is degraded. These attributes do not contribute to the blurriness of the output because the HR corresponding to the LR has a one-to-one correlation when using a conventional face SR framework. In contrast, the stochastic attributes refer to the attributes that have partially or completely lost their information and thus are subject to probabilistic prediction from the LR image. Depending on the degree of information loss, the stochastic attributes can be divided into two subcategories. In the partially alive case shown in Figure \ref{fig:Information Flow in FSR}, the attributes such as the eyebrow shape can be inferred probabilistically in the form of a Gaussian from the LR image. In the fully lost case, the probability of the attributes can be assumed to follow a uniform distribution because the attributes such as a wrinkle on a cheek cannot be inferred.

Specifically, for the stochastic attributes, several HR images exist that correspond to the LR image, and these aspects are accepted by the network as the uncertainty. In conventional face SR, there was no means of removing this uncertainty in the decoding phase which makes the final result image from the extracted information. The network thus reflects all the possibilities in the single output image, resulting in blurry shapes such as the eyebrows, as shown in Figure \ref{fig:Blurry Outputs}.

To overcome this limitation, this paper proposes a novel face SR framework that can remove the blur effect caused by the uncertainty from the stochastic attributes. First, the information of the LR image is separated into deterministic and stochastic attributes by using the error encoding framework \cite{EEN}. The extracted stochastic attributes are encoded into the latent space in the form of a multivariate Gaussian distribution by using a Residual Encoding Network (REN). The stochastic attributes can be selected by sampling from the distribution and later added to a middle feature map of the face SR network. When we get the stochastic attributes in training phase, it is necessary to utilize the HR image in the proposed framework. However, it is not possible to show the HR image in the inference phase. To solve this problem, an Input Conditional Attribute Predictor (ICAP) is proposed, which requires only an LR image as the input to infer the stochastic attributes. This network is forced to imitate the output values of the REN so that the network can predict the partially alive stochastic attributes from the LR image. We conduct experiments on UTKFace and CelebA datasets using the proposed method, and as a result, demonstrate the superiority of the method.

The main contributions of our work are three-fold:
\begin{itemize}
    \item To the best of our knowledge, this is the first face SR framework that can eliminate the blurry effect caused by the uncertainty from the stochastic attributes.
    \item The deterministic and stochastic attributes is separated from the LR face images in unsupervised manner.
    \item Using the proposed ICAP, the partially alive stochastic attributes can be inferred by considering only the LR input, thereby eliminating the dependence of the framework on the ground-truth HR image.
\end{itemize}


\section{Related Works}
In the face SR domain, there are many prior works, which tries to make a clear SR output. The recent approaches can be divided into two categories. 
The first one involves using the the facial prior knowledge in the learning process. In an existing study, Super-FAN \cite{Super-FAN} proposed the framework that addresses face SR and alignment simultaneously to improve the SR performance. In the FSRNet \cite{FSRNet}, the facial boundary was attempted to be clarified by using a geometry prior such as parsing maps and facial landmarks. In the RAAN \cite{RAAN-GAN}, both the pixel level and semantic level facial priors were utilized. Most of these methods can be interpreted as an effort to reduce the width of the Gaussian distribution illustrated in Figure \ref{fig:Information Flow in FSR}. Such techniques can effectively improve the results; however, the outputs are still blurry because the stochastic attributes still remain in the Gaussian form.

The second approach utilizes adversarial learning framework, inspired by Generative Adversarial Network (GAN) \cite{GAN}. It is a consensus that GAN-based models can generate photo realistic images. There are a lot of papers such as SRGAN \cite{SRResNet} which utilize adversarial loss with pixel-wise loss in general SR problem. Furthermore, in the face SR domain, the approaches involving URDGN \cite{URDGN}, FSRGAN \cite{FSRNet}, and RAAN-GAN \cite{RAAN-GAN} adopted the adversarial training method, yielding visually plausible results. However, quantitatively, all these approaches suffer from a low peak signal-to-noise ratio (PSNR) and structural similarity (SSIM). This means that the resulting image is considerably different from the correct HR image. This is because GAN-based methods attempt to create the output images that have the same distribution as that of the HR images, and the focus is not on restoring the correct HR corresponding to the LR image. In the general SR domain \cite{GeneralSR1,GeneralSR2,GeneralSR3,GeneralSR4}, which yields just visually plausible results is sufficient. However, the finding is not consistent with the goal of the face SR, that a person's identity should be preserved. Therefore, in this work, we attempt to produce a clear output without using the GAN-based method of changing a person's identity.

\section{Method}
\label{ch:3}
The objective here is to reduce the blurry effect caused by stochastic attributes in the training phase. The proposed approach can be divided into two stages. First, the information contained in an LR image is separated into deterministic and stochastic attributes by utilizing the Error Encoding Network \cite{EEN}. The stochastic attributes from the residual image is encoded to a low-dimensional latent space in the form of a multivariate Gaussian distribution. Finally, we train another network that can predict the partially alive stochastic attributes directly from the LR image without using the ground-truth HR image in the inference.

\begin{figure}[t]
    \centerline{\includegraphics[width=126mm]{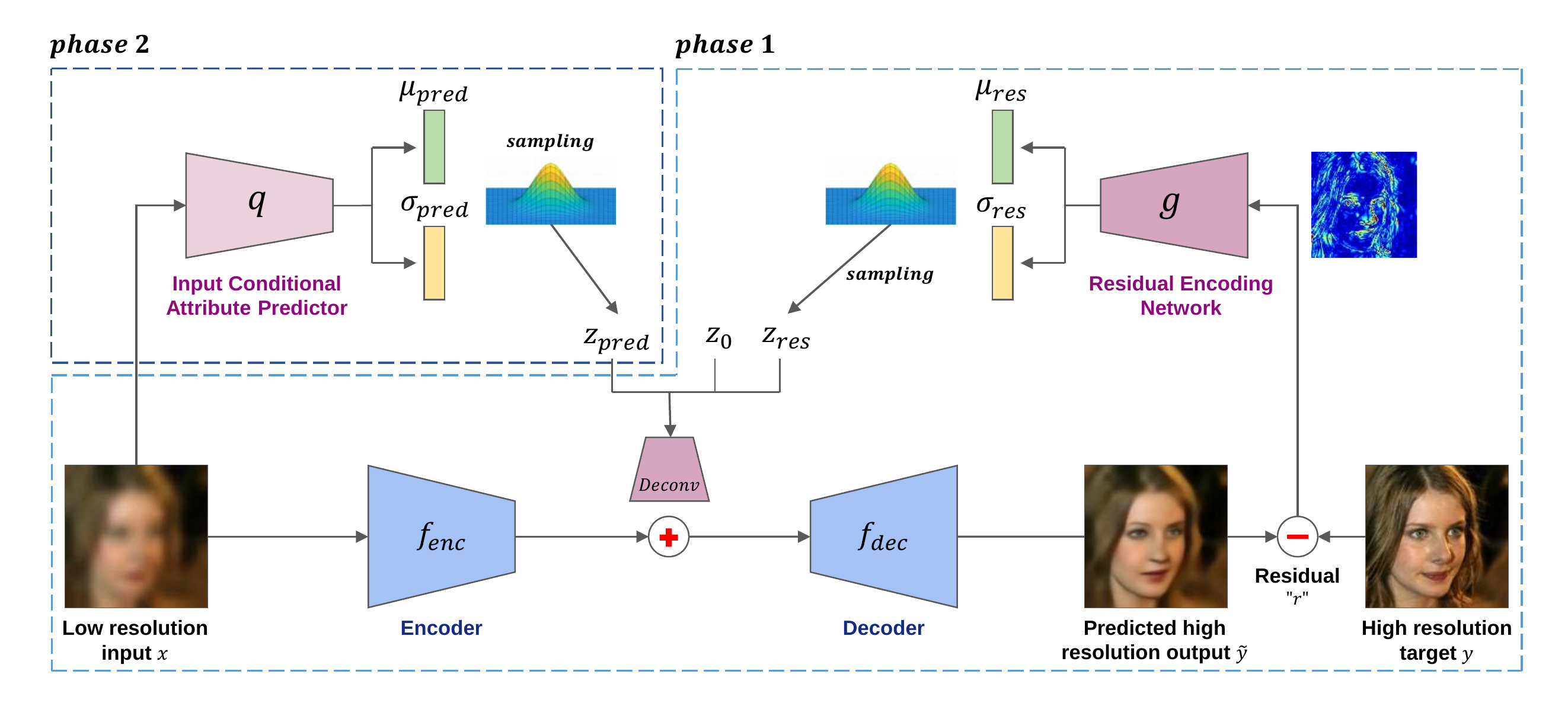}}
    \caption[Network structure of the proposed method]{
    Network structure of the proposed method.
    } \label{fig:Network Structure of Proposed Method}
\end{figure}

\subsection{Network Architecture}
As shown in Figure \ref{fig:Network Structure of Proposed Method}, the proposed networks mainly consist of three parts: \textit{SR Encoder-Decoder}, \textit{Residual Encoding Network}, and \textit{Input Conditional Attribute Predictor}. We now present each part of the network as step by step. For training phase, we use a given training set $\{\mathbf{x}^{(i)}, \mathbf{y}^{(i)}\}_{i=1}^N$, where $N$ is the number of training images, $\mathbf{x}^{(i)}$ is the LR image, and $\mathbf{y}^{(i)}$ is the ground-truth HR image corresponding to $\mathbf{x}^{(i)}$.

\subsubsection{SR Encoder-Decoder}
The SR Encoder-Decoder, which is denoted as $f_\theta$ with parameters $\theta$, is the main face SR network in the proposed framework. This makes a $128\times128$ SR image $\tilde{\mathbf{y}}$ from a given $128\times128$ bicubic upsampled LR image $\mathbf{x}$. The network is composed of the encoder network $f_{enc}$ that encodes the information of the LR image, and the decoder network $f_{dec}$ that creates a final output image from the encoded feature map. Additionally, there is a deconvolution layer to accept the external information, which comes from a latent vector $\vect{z}$, and adds this to the output feature map of the encoder. The deconvolutional layer allows the SR Encoder-Decoder to create an output with a more detailed facial representation.

The SR Encoder-Decoder can produce outputs of different quality according to the sampled $\vect{z}$ vectors. If a latent vector $\vect{z}$ is set as a zero vector $\vect{z}_{\vect{0}}$, the network generates a blurry output as a predicted HR output $\tilde{\mathbf{y}}$ without the external information. In this case, the network infers the HR target $\mathbf{y}$ as much as possible without using the external information. 

Consequently, the information mainly used in an LR image is related to the deterministic attributes, which can be clearly predicted the HR's attributes from an LR image. This output image known as $\tilde{\mathbf{y}}_{\mathbf{d}}$ is obtained using the following equation:
\begin{equation}
\begin{aligned}
  \tilde{\mathbf{y}}_{\mathbf{d}} = \mathbf{f}_{\theta}(\mathbf{x},  \mathbf{z}=\vect{z}_{\vect{0}}).
  \end{aligned}
  \label{eq:3.1}
\end{equation}

If a latent vector $\vect{z}$ is set as $\vect{z}_{res}$, which includes the external information from the Residual Encoding Network, the network generates a high-quality output as the predicted HR output $\mathbf{y}$ by using the information obtained from both LR input and $\vect{z}_{res}$. In this case, the network generates the output by using the latent vector $\vect{z}_{res}$, which includes the stochastic attributes. The output image in this case, $\tilde{\mathbf{y}}_{\mathbf{s}}$, is obtained using the following equation:
\begin{equation}
\begin{aligned}
  \tilde{\mathbf{y}}_{\mathbf{s}} = \mathbf{f}_{\theta}(\mathbf{x},  \mathbf{z}=\vect{z}_{res}).
  \end{aligned}
  \label{eq:3.2}
\end{equation}

We call each process of generating $\tilde{\mathbf{y}}_{\mathbf{d}}$ and $\tilde{\mathbf{y}}_{\mathbf{s}}$ as a \textit{deterministic path} and \textit{stochastic path}, respectively. First, a low-quality image $\tilde{\mathbf{y}}_{\mathbf{d}}$ is obtained through the \textit{deterministic path} without reference to any external information. Next, a high-quality image $\tilde{\mathbf{y}}_{\mathbf{s}}$ is created from an input LR image, referring to $\vect{z}_{res}$ obtained from the residual between $\mathbf{y}$ and $\tilde{\mathbf{y}}_{\mathbf{d}}$.

\subsubsection{Residual Encoding Network}
The Residual Encoding Network (REN), which is denoted as $g_\phi$ with parameters $\phi$, is used to encodes the stochastic attributes into the low-dimensional latent space. The network uses the residual image $\mathbf{r}$, which is obtained using the following equation:
\begin{equation}
\begin{aligned}
  \mathbf{r} = \mathbf{y} - \tilde{\mathbf{y}}_{\mathbf{d}}.
  \end{aligned}
  \label{eq:3.3}
\end{equation}
The residual of the blurry image $\tilde{\mathbf{y}}_{\mathbf{d}}$, which is a result of the failure to perform the restoration in the \textit{deterministic path}, is considered as the stochastic attributes. When the network encodes the stochastic attributes into latent space, the encoding result is in the form of a multivariate Gaussian distribution parameterized by the mean $\vect{\mu}_{res}$ and standard deviation $\vect{\sigma}_{res}$. We sample from the posterior $\vect{z}_{res} \sim g_\phi(\cdot|\mathbf{r})$ using $\vect{z}_{res} = \vect{\mu}_{res} + \vect{\sigma}_{res} \odot \vect{\epsilon}$, where $\vect{\epsilon}$ is drawn from $\mathcal{N}(\mathbf{0}, \mathbf{I})$ and $\odot$ denotes the element-wise multiplication.

Consequently, the closest attribute to that of the HR image is the mean vector of the distribution, and the other attributes that are different from those of the HR image have a lower probability. Furthermore, by selecting only one of the possible attributes through sampling, the blurry effect in the output images, which arises because of the uncertainty caused by multiple possibilities, can be eliminated.

We train the networks described so far as follows: First, the SR encoder-decoder is trained to predict the HR as much as possible from the LR with the latent vector $\vect{z}$ set to zero to ensure that the deterministic information can be fully used without any additional information. The SR Encoder-Decoder is trained using the pixel-wise Mean Squared Error (MSE) function as follows:
\begin{equation}
\begin{aligned}
  \mathcal{L}_{\mathbf{d}}(\theta) = & \frac{1}{N}\sum_{i=1}^{N}\|\mathbf{y}^{(i)} - \mathbf{f}_{\theta}(\mathbf{x}^{(i)}, \mathbf{z}=\vect{z}_{\vect{0}})\|_{2}^{2}.
  \end{aligned}
  \label{eq:3.6}
\end{equation}

To ensure that the REN successfully encodes the stochastic information and the SR Encoder-Decoder reflects the encoded vector into the predicted output image, the pixel-wise L1 loss function was adopted:
\begin{equation}
\begin{aligned}
  \mathcal{L}_{\mathbf{s}}(\theta,\phi) = & \frac{1}{N}\sum_{i=1}^{N}\|\mathbf{y}^{(i)} - \mathbf{f}_{\theta}(\mathbf{x}^{(i)}, \mathbf{z}=\vect{z}_{res})\|_{1}.
  \end{aligned}
  \label{eq:3.7}
\end{equation}
In this case, we adopt the L1 loss to better preserve the stochastic attributes by imposing stronger constraints. Furthermore, the parameters of the SR Encoder-Decoder for the \textit{deterministic path} and \textit{stochastic path} is shared and trained jointly using the following loss function:
\begin{equation}
\begin{aligned}
  \mathcal{L}_{phase 1}(\theta,\phi) = \mathcal{L}_{\mathbf{d}}(\theta) + \lambda \cdot \mathcal{L}_{\mathbf{s}}(\theta,\phi)
  \end{aligned}
  \label{eq:3.8}
\end{equation}
We constrain the values of $\vect{\mu}_{pred}$ and $\vect{\sigma}_{pred}^2$ to the specific range instead of making the distribution of $\vect{z}_{res}$ similar to prior distribution due to the posterior collapse.

\subsubsection{Input Conditional Attribute Predictor}
We divided the information contained in an LR image into deterministic and stochastic attributes and successfully reflected the sampled latent vector in an SR output. However, the HR images must be used to obtain the stochastic attributes in the current framework. This is not consistent with the our goal to obtain a clear predicted output image using only a single LR image. Therefore, we propose the Input Conditional Attribute Predictor (ICAP), which is denoted as $q_\omega$ with parameters $\omega$. This predictor can be used to immediately predict the partially alive stochastic attributes from the LR image in the inference phase. The ICAP network outputs the stochastic attributes $\vect{z}_{pred}$ reparameterized by $\vect{\mu}_{pred}$ and $\vect{\sigma}_{pred}$ similar to the REN. However, because this network receives only an LR image as the input, the network is trained using the loss function, as follows:
\begin{equation}
\mathcal{L}_{phase 2}(\omega) = D_\text{KL} \big(q_\omega(\vect{z}_{pred}|\mathbf{x})||g_\phi(\vect{z}_{res}|\mathbf{r})\big).
\label{eq:3.10}
\end{equation}
Consequently, the network can infer the characteristics of HR from the LR image by making the output distributions of the ICAP and REN closer.

\subsection{Training Procedure}
This section describes the whole training procedure of the proposed method. As shown in Algorithm \ref{alg:alg1}, the training procedure can be divided into two phases: \textit{phase1} and \textit{phase2}. In \textit{phase1}, both the SR Encoder-Decoder and REN are trained by directly referring to the ground-truth HR images when generating the predicted HR images. In this process, we update the parameters $\theta$ and $\phi$ with the loss function described in Equation \ref{eq:3.8} until the networks are sufficiently converged. In \textit{phase2}, the ICAP is trained using the ground-truth $\vect{\mu}_{res}$ and $\vect{\sigma}_{res}$ obtained from the \textit{phase1}.
Subsequently, the parameters $\omega$ of the ICAP is updated using the KL divergence loss function described in Equation \ref{eq:3.10} to ensure that the output distribution of the ICAP is similar to the ground-truth distribution. While training the ICAP, the parameters $\theta$ and $\phi$ of the SR Encoder-Decoder and REN are frozen.

\begin{algorithm}[t]
  \caption{Training Procedure of the Proposed Method}
  {\bf Data:}
  Given training set $\{\mathbf{x}^{(i)}, \mathbf{y}^{(i)}\}_{i=1}^N$, where $N$ is the number of training images, $\mathbf{x}^{(i)}$ is the LR image, and $\mathbf{y}^{(i)}$ is the ground-truth HR image corresponding to $\mathbf{x}^{(i)}$.
  
  {\bf Initialize:}
  All the settings include the learning rates, optimization method, training iteration, \textit{etc.}

  {\bf Output:}
  Proposed Network
  \begin{algorithmic}
  \label{alg:alg1}
  \WHILE{\textit{not converged}}
  \STATE Update the parameter sets $\theta$ and $\phi$ $\gets$ minimize loss $\mathcal{L}_{phase 1}$
  \ENDWHILE
  \WHILE{\textit{not converged}}
  \STATE Update the parameter set $\omega$ $\gets$ minimize loss $\mathcal{L}_{phase 2}$
  \ENDWHILE
  \end{algorithmic}
\end{algorithm}

\section{Experiments}
Extensive experiments were conducted to verify the effectiveness of the proposed method. In this section, first, we explain the implementation details of the conducted experiments. To demonstrate the quantitative and qualitative superiority of the proposed method compared to the state-of-the-art approaches, two kinds of experiments were performed, and the results are described herein. Furthermore, ablation studies were performed to validate the network performance in terms of the stochastic attribute modeling.

\subsection{Implementation Details}
Experiments were conducted on two datasets: UTKFace \cite{UTKFace} and, CelebA \cite{CelebA}. For training, we use the randomly chosen $23,408$ images for UTKFace dataset and $28,970$ images for CelebA dataset without any augmentation methods. For evaluation, 300 images are randomly chosen from each dataset. The proposed model is trained using the Adam optimizer with an initial learning rate of $2.5$ $\times$ $10^{-4}$ for \textit{phase1} and $0.005$ for \textit{phase2}. We constrain the values of $\vect{\mu}_{pred}$ and log of $\vect{\sigma}_{pred}^2$ to the range $\mathbf{[-1,1]}$ to make a compressed distribution in the latent space for the training in \textit{phase1}. All the input LR images are enlarged by bicubic upsampling before using because the proposed model has a fixed input size $128$ $\times$ $128$, same as that of the output.

\subsection{Residual Encoding Results}
In this section, before validating the final results of generating the HR image using only an LR image, we ensure that the network successfully encodes the stochastic attributes and reflects the value in the final results. To this end, we generate the SR results using both the LR and HR images by employing both the SR Encoder-Decoder and REN.
\begin{figure}[t!]
    \centerline{\includegraphics[width=\textwidth]{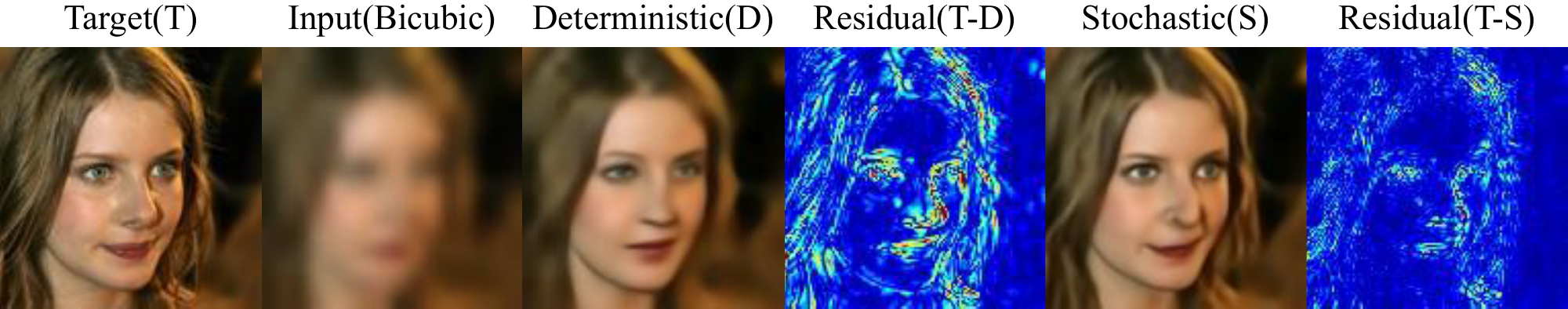}}
    \caption[Visual result of residual encoding]{
    Visual result of residual encoding.
    } \label{fig:Residual Encoding Result}
\end{figure}
Figure \ref{fig:Residual Encoding Result} shows the visual result of the residual encoding. \textit{Target(T)} and \textit{Input(Bicubic)} denote the ground-truth HR image and degraded LR image that the network uses to make the SR output, respectively. \textit{Deterministic(D)} is the result of using only the LR image with the SR Encoder-Decoder with $\vect{z}=\vect{0}$. In this case, the network generates the result using only the deterministic information. The \textit{Residual(T-D)} can be obtained using the difference between \textit{T} and \textit{D} determined previously, which contains the information that is difficult to predict from \textit{Input(Bicubic)} under the condition of $\vect{z}=\vect{0}$. The REN encodes this \textit{Residual(T-D)} using a multivariate Gaussian distribution. \textit{Stochastic(S)} is the result of reflecting the specific vector sampled from the encoded distribution. As shown in \textit{Residual(T-S)}, which represents the difference between \textit{T} and \textit{S}, we can see that \textit{S} is about the same as \textit{T}. These findings confirm that the proposed network can separate the stochastic information from the LR image and encode it, and this information is successfully reflected into the final results.

\begin{table}[t!]
\caption[Quantitative comparisons with PSNR/SSIM]{
Quantitative comparisons with PSNR/SSIM. The \textcolor{red}{red}/\textcolor{blue}{blue} colors indicate the highest/second-highest performance.}
\begin{center}
\footnotesize
\resizebox{\textwidth}{!}{\begin{tabular} {|c|c|c|c|c|c|c|c|}
\hline
Scale factor & Bicubic & SRCNN & VDSR & SRResNet & URDGN & Wavelet-SRNet & Ours\\
\hline\hline
$4$ &  $28.04$/$0.86$ & $31.69$/$0.91$ & \textcolor{blue}{$33.44$}/\textcolor{blue}{$0.93$} & $31.48$/$0.90$ & $30.79$/$0.90$ & $31.88$/$0.91$ & \textcolor{red}{$33.52$}/\textcolor{red}{$0.94$} \\
$8$ & $23.11$/$0.71$ & $25.21$/$0.76$ & \textcolor{blue}{$26.73$}/$0.81$ & $26.51$/$0.80$ & $25.52$/$0.77$ & $26.66$/\textcolor{blue}{$0.82$} & \textcolor{red}{$27.86$}/\textcolor{red}{$0.85$} \\
$16$ & $19.51$/$0.59$ & $20.88$/$0.62$ & $21.75$/$0.65$ & \textcolor{blue}{$22.73$}/\textcolor{blue}{$0.70$} & $21.82$/$0.66$ & $22.53$/$0.66$ & \textcolor{red}{$23.44$}/\textcolor{red}{$0.74$} \\
\hline\hline
$4$  & $24.88$/$0.76$ & $27.32$/$0.83$ & \textcolor{blue}{$29.03$}/\textcolor{blue}{$0.88$} & $28.00$/$0.84$ & $27.09$/$0.83$ & $28.35$/$0.86$ & \textcolor{red}{$29.24$}/\textcolor{red}{$0.88$} \\
$8$  & $21.15$/$0.60$ & $23.10$/$0.67$ & \textcolor{blue}{$24.38$}/\textcolor{blue}{$0.73$} & $24.05$/$0.72$ & $23.49$/$0.69$ & $23.80$/$0.72$ & \textcolor{red}{$24.91$}/\textcolor{red}{$0.76$} \\
$16$  & $17.95$/$0.48$ & $19.47$/$0.52$ & $20.17$/$0.56$ & \textcolor{blue}{$20.79$}/\textcolor{blue}{$0.60$} & $20.29$/$0.57$ & $20.34$/$0.56$ & \textcolor{red}{$21.33$}/\textcolor{red}{$0.64$} \\
\hline
\end{tabular}}
\end{center}
\label{tab:Benchmark for UTKFace, CelebA}
\end{table}

\subsection{Comparison with State-of-the-Art Methods}
In this section, we compare our final framework, which makes a predicted HR output from only a single LR image using both the SR Encoder-Decoder and the ICAP with state-of-the-art SR methods which do not use facial priors, such as the SRCNN \cite{SRCNN}, VDSR \cite{VDSR}, SRResNet \cite{SRResNet}, URDGN \cite{URDGN}, and Wavelet-SRNet \cite{Wavelet-SRNet}.

\begin{figure}[t!]
    \centerline{\includegraphics[width=140mm]{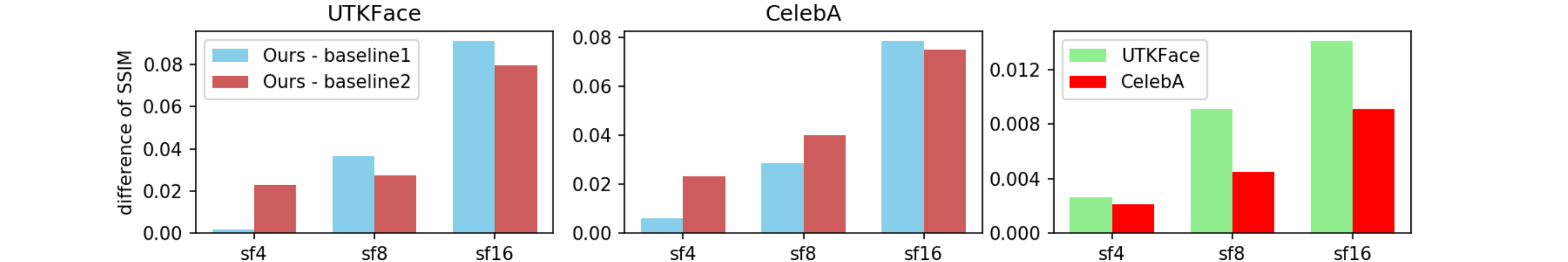}}
    \caption[Performance difference with the scale factor]{
    \textbf{Left}: The difference of SSIM between the proposed method and two baselines on UTKFace. The baseline1 is VDSR \cite{VDSR}, and the baseline2 is Wavelet-SRNet \cite{Wavelet-SRNet}. \textbf{Middle}: The difference of SSIM between the proposed method and two baselines on CelebA. \textbf{Right}: The difference of SSIM between the proposed method and our deterministic baseline.
    } \label{fig:Strength for high sf}
\end{figure}

First, a quantitative analysis is performed. Table \ref{tab:Benchmark for UTKFace, CelebA} presents the quantitative results with scale factor of 4, 8, and 16 for datasets. It can be noted that the proposed method significantly outperforms the state-of-the-art methods on both the datasets and all the scale factors in terms of both the PSNR and SSIM. Additionally, Figure \ref{fig:Strength for high sf} shows that a lower resolution of the input image corresponds to a higher performance. This means that the proposed method can better restore the image as the proportion of the stochastic information increases.

\begin{figure}[t!]
    \centerline{\includegraphics[width=122mm]{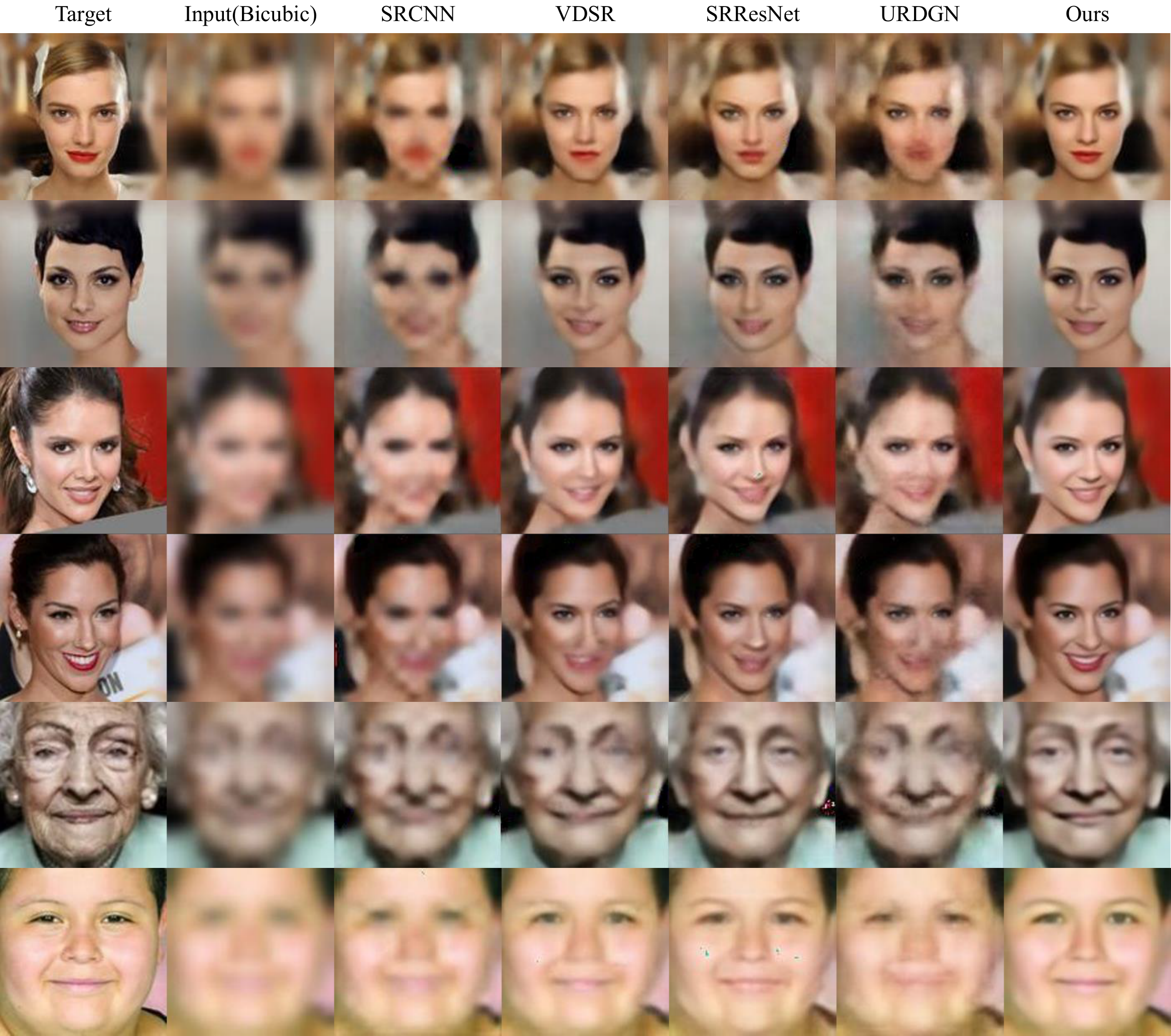}}
    \caption[Qualitative comparisons for a scale factor of 8]{
    Qualitative comparisons for a scale factor of 8. The images shown in the top four rows are derived from CelebA, and the other images are from the UTKFace dataset.
    } \label{fig:Qualitative comparisons}
\end{figure}

\begin{figure}[p]
    \centerline{\includegraphics[width=122mm]{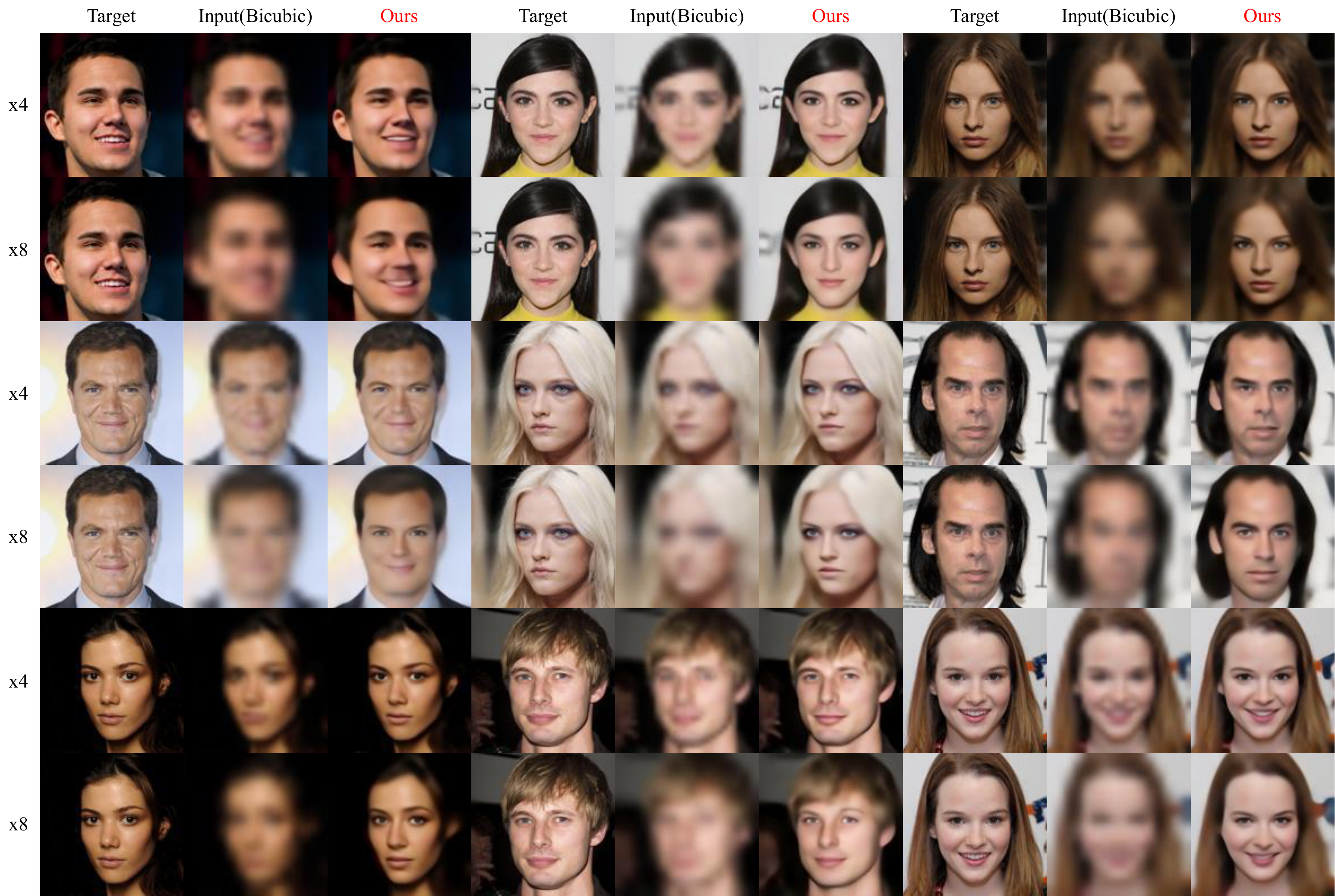}}
    \caption[Qualitative results for the CelebA dataset for scale factors 8 - 1]{
    Qualitative results for the CelebA dataset for scale factors 4 and 8 - set1.
    } \label{fig:more results 1}
\end{figure}

\begin{figure}[p]
    \centerline{\includegraphics[width=122mm]{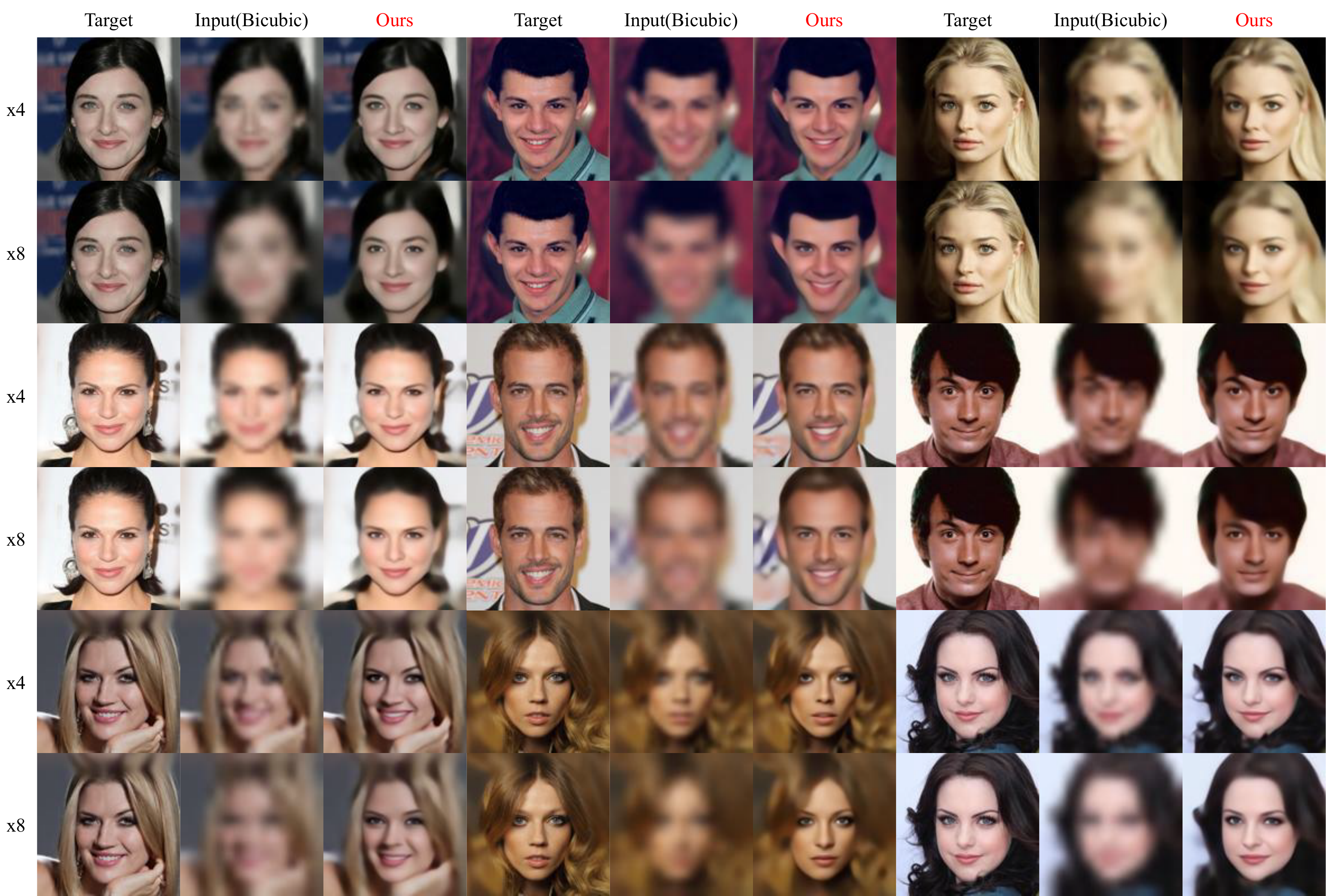}}
    \caption[Qualitative results on CelebA dataset for scale factor 8 - 2]{
    Qualitative results for the CelebA dataset for scale factors 4 and 8 - set2.
    } \label{fig:more results 2}
\end{figure}

Figure \ref{fig:Qualitative comparisons} shows the qualitative comparison of the proposed method for a scale factor of 8. It can be noted that the proposed method generates relatively sharper shapes and edges for the important attributes in a person's face such as the eyebrow, lips, and nose. This considerably helps restoration of a person's impression, and therefore, the proposed method can retain the identity of the individual. More qualitative results pertaining to the scale factors 4 and 8 are shown in Figures \ref{fig:more results 1} and \ref{fig:more results 2}, respectively.

\subsection{Ablation Study}
The objective in this study is to reduce the eliminate the inherent uncertainty in the existing face SR networks by extracting the stochastic attributes and modeling them as a latent space in the form of a probability distribution. Although we have demonstrated the excellent performance of the proposed method, as described in the previous subsection, this alone is not sufficient to confirm that the encoded result is stochastically meaningful. Therefore, in this part, we described the evaluation of our method regarding the encoding of the stochastic information into meaningful latent spaces through various ablation studies.

\subsubsection{Multiple Sampling Results}
We encode the stochastic attributes into a multivariate Gaussian distribution, assuming that the attributes closest to the HR image should be a mean vector having the highest probability in the distribution. To demonstrate the effectiveness of the proposed method, when sampling the latent vector from the distribution parameterized by $\vect{\mu}_{pred}$ and $\vect{\sigma}_{pred}$, the mean vector should yield a result closest to that for the HR image such that it can be inferred from the LR image. To this end, the following experiment is performed.

\begin{figure}[t!]
    \centerline{\includegraphics[width=110mm]{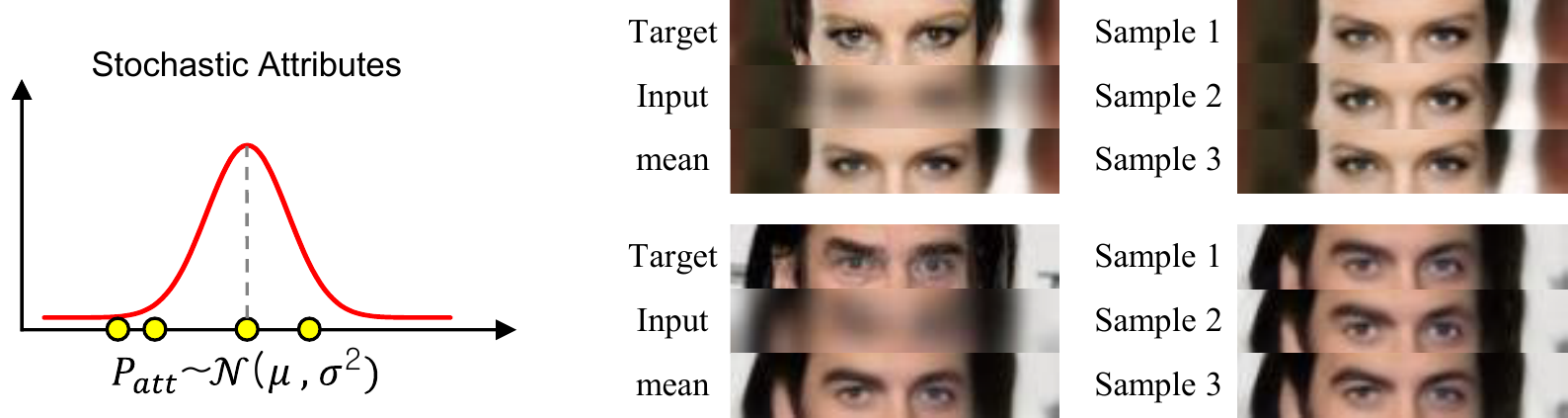}}
    \caption[Multiple sampling results with $\vect{\mu}_{pred}$ and $\vect{\sigma}_{pred}$]{
    Multiple sampling results with $\vect{\mu}_{pred}$ and $\vect{\sigma}_{pred}$. \textit{mean} denotes the sample pertaining to $(\vect{\mu},\vect{\sigma})=(\vect{\mu}_{pred},\vect{0})$ and \textit{Samples 1} through \textit{3} denote the three sampling results from $(\vect{\mu},\vect{\sigma})=(\vect{\mu}_{pred},\vect{\sigma}_{pred})$.
    } \label{fig:Multiple sampling results}
\end{figure}

We predict $\vect{\mu}_{pred}$ and $\vect{\sigma}_{pred}$ from an LR image by using the ICAP. Next, we use $\vect{\mu}_{pred}$ as the closest attribute to the ground-truth and obtain the three other latent vectors by sampling from the distribution. Finally, the final SR results, named \textit{mean}, \textit{Sample 1}, \textit{Sample 2}, and \textit{Sample 3}, are generated by adding the latent vectors to the middle feature map of SR Encoder-Decoder. The results are shown in Figure \ref{fig:Multiple sampling results}. It can be observed that the eyebrow shape of \textit{mean} is closer to the HR image than that for the other samples far from the center of the probability distribution. Therefore, the proposed method can be considered to be effective.

Furthermore, we determine the best PSNR/SSIM performance using only the best results after sampling multiple result images from the predicted distribution. The procedure of obtaining the samples is the same as that in the previous experiments. The results of the experiment are presented in Table \ref{tab:Multiple sampling}. \textit{w/o sampling} in the table corresponds to the case of in which $\vect{\mu}_{pred}$ is used as the latent vector, and \textit{w/ sampling} is the case in which the distribution parameterized by $\vect{\mu}_{pred}$ and $\vect{\sigma}_{pred}$ is sampled. \textit{n} refers to the number of samplings. The results show that a higher number of samples correspond to a higher PSNR/SSIM for each setting of the dataset and scale factor.

\begin{table}[t!]
\caption[Highest PSNR/SSIM performances of multiple sampling results with $\vect{\mu}_{pred}$ and $\vect{\sigma}_{pred}$]{
Highest PSNR/SSIM performances of multiple sampling results with $\vect{\mu}_{pred}$ and $\vect{\sigma}_{pred}$.
}
\begin{center}
\footnotesize
\resizebox{\textwidth}{!}{\begin{tabular} {|ccccccccc|}
\hline
&&& w/o sampling && \multicolumn{3}{c}{w/ sampling} & \\
\cline{4-4} \cline{6-8}
& Dataset & Scale factor & mean && n=10 & n=100 & n=1000 & \\
\hline\hline
& UTKFace & $4$ & $33.5196$/$0.9362$ && $33.5262$/$0.9362$ & $33.5291$/$0.9363$ & $33.5314$/$0.9363$ & \\
& CelebA & $4$ & $29.2391$/$0.8816$ && $29.2744$/$0.8822$ & $29.2949$/$0.8826$ & $29.3100$/$0.8828$ & \\
\hline
& UTKFace & $8$ & $27.8597$/$0.8457$ && $27.9778$/$0.8478$ & $28.1203$/$0.8507$ & $28.2345$/$0.8528$ & \\
& CelebA & $8$ & $24.9094$/$0.7574$ && $25.0080$/$0.7600$ & $25.0905$/$0.7627$ & $25.1572$/$0.7646$ & \\
\hline
& UTKFace & $16$ & $23.4414$/$0.7384$ && $23.5632$/$0.7433$ & $23.7190$/$0.7486$ & $23.8462$/$0.7530$ & \\
& CelebA & $16$ & $21.3295$/$0.6339$ && $21.4404$/$0.6394$ & $21.5698$/$0.6454$ & $21.6533$/$0.6496$ & \\
\hline
\end{tabular}}
\end{center}
\label{tab:Multiple sampling}
\end{table}

In Section \ref{ch:1}, we classified the stochastic information contained in the LR images as partially alive and fully lost attributes. Although the ICAP can predict the stochastic attributes, this approach is only applicable to the partially alive case. The fully lost information can still not be determined. Therefore, the results presented in Table \ref{tab:Multiple sampling} can be interpreted as follows: When multiple sampling is performed, the fully lost attributes are sometimes accidentally matched while maintaining the prediction of the partially alive attributes, which is reflected in the performance increase.

\subsubsection{Traversing Latent Space}
If the encoded attributes exist in a meaningful latent space, significant and continuous changes in the attributes must occur when sliding the latent vector. To confirm this, the latent traversal experiment was performed.

We first obtain $\vect{\mu}_{pred}$ and $\vect{\sigma}_{pred}$ from an LR image using the ICAP as in previous experiments. Then, two latent vectors are obtained from the distribution by sampling. These two vectors denote the start and end points in the latent space, denoted as $\vect{z}_{start}$ and $\vect{z}_{end}$, respectively. And then, the other latent vectors denoted as $\vect{z}_{inter}$ are obtained by the convex combination of $\vect{z}_{start}$ and $\vect{z}_{end}$ following the expression:

\begin{equation}
\begin{aligned}
  \vect{z}_{inter} = \vect{z}_{start} + \alpha \cdot (\vect{z}_{end} - \vect{z}_{start}).
  \end{aligned}
  \label{eq:Latent traversal}
\end{equation}

\begin{figure}[t!]
    \centerline{\includegraphics[width=\textwidth]{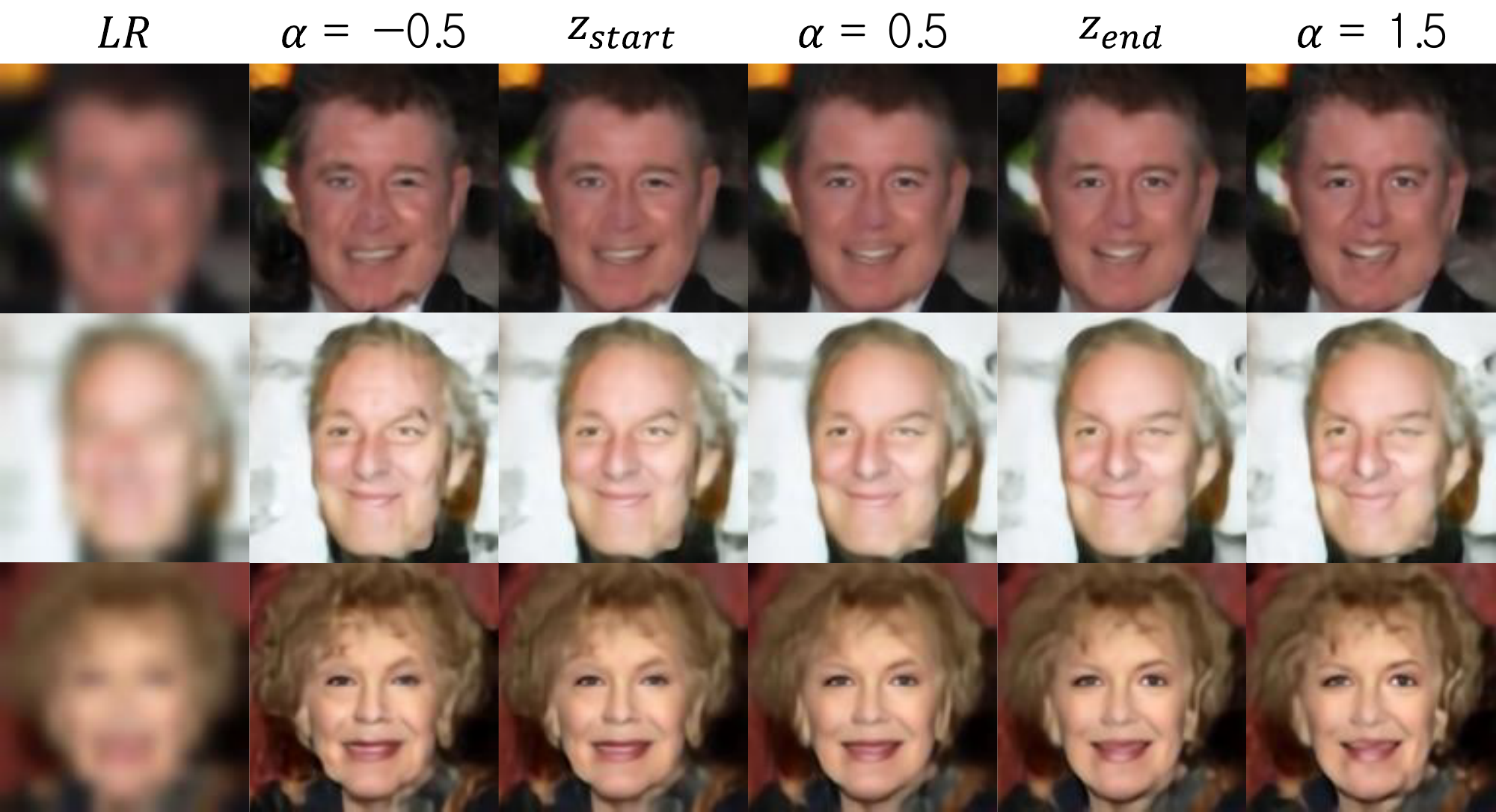}}
    \caption[Manipulating latent variables]{
    Manipulating latent variables
    } \label{fig:Traversing latent space}
\end{figure}

Then, we generate the final SR results using the latent vectors. The results are shown in Figure \ref{fig:Traversing latent space}. When the latent vector is changed in a particular direction, the resulting image changes continuously. This aspect confirms that the encoded stochastic attributes exist in a meaningful latent space.
	
\section{Conclusions}
This paper proposes a novel method for face SR, which can remove the uncertainty caused by the stochastic attributes in the learning process. The key components of the proposed method are the RENs, which can encode the stochastic attributes into compressed multivariate Gaussian distribution, and the ICAP, which can predict the partially alive stochastic attributes from the LR image. The ablation study, which involves multiple sampling from the predicted attribute distribution and latent space traversal experiments, confirmed that the proposed method can successfully perform stochastic modeling for stochastic attributes. Furthermore, the results of extensive experiments performed on CelebA and UTKFace demonstrates that the proposed method outperforms the state-of-the-art approaches, both qualitatively and quantitatively. In this work, the facial priors such as facial landmarks and parsing maps were not used to focus on eliminating the uncertainty. Therefore, in future work, we aim to combine these priors with the proposed method to achieve better performance.

\clearpage
%
%
\bibliographystyle{splncs04}
\bibliography{egbib}
\clearpage

\section{Appendix}
This supplementary material provides additional details as the following:
\begin{itemize}
    \item Final framework of the proposed method for the inference phase.
    \item The network structure for each components of the proposed method.
\end{itemize}

\subsection{Framework for the Inference Phase}
\begin{figure}[!]
    \centerline{\includegraphics[width=\textwidth]{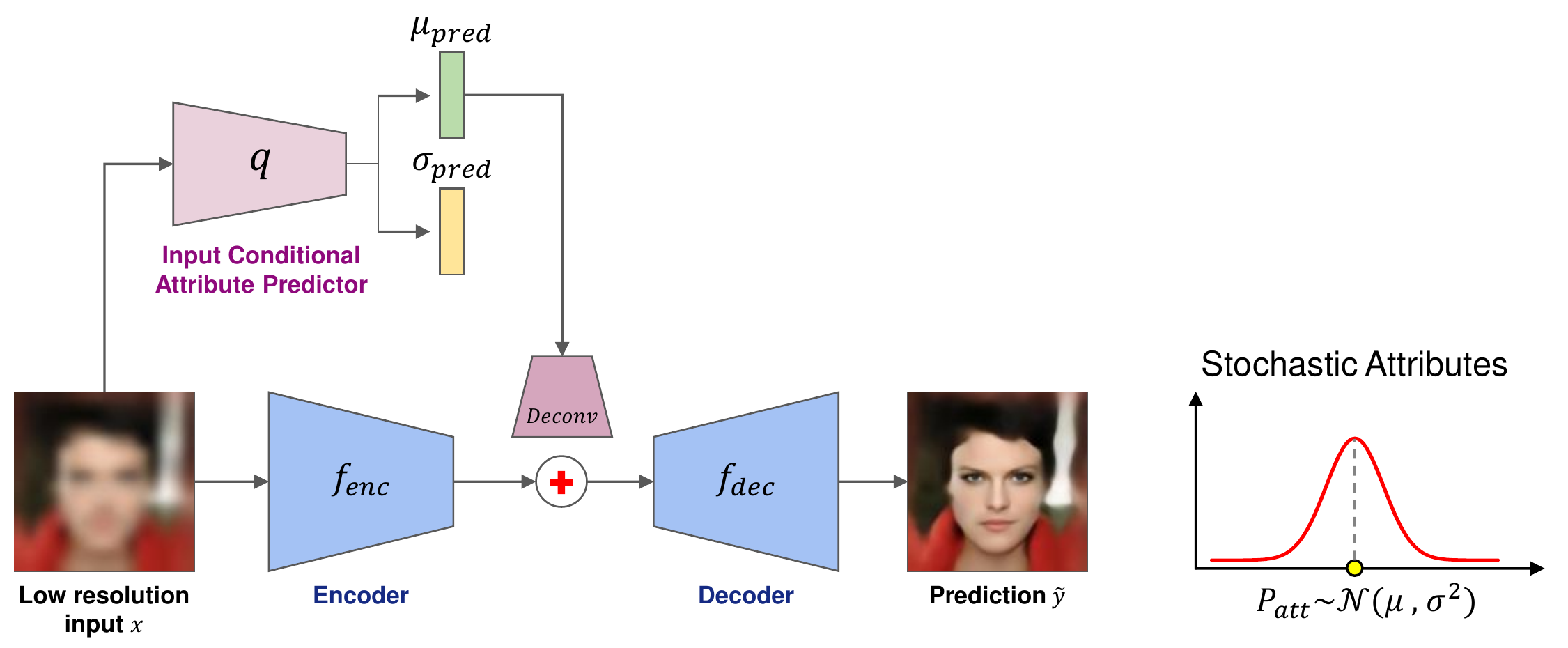}}
    \caption[Framework of the proposed method for the inference phase]{
    Framework of the proposed method for the inference phase
    } \label{fig:Framework_Inference_with_Image}
\end{figure}

The final framework, which makes a predicted HR output from only a single LR image in the inference phase is described as Figure \ref{fig:Framework_Inference_with_Image}.

The Input Conditional Attribute Predictor predicts the distribution of stochastic attributes from the LR image. Then the mean vector, which has the highest probability value in the distribution is send to the feature maps in the middle of the SR Encoder-Decoder. Finally, the SR Encoder-Decoder generates the HR prediction using both the deterministic attributes and stochastic attributes.
\clearpage

\subsection{Network Structure}
Tables \ref{tab:SREncoderDecoder}, \ref{tab:ICAP} and \ref{tab:REN} present the network structure of the SR Encoder-Decoder, Input Conditional Attribute Predictor and Residual Encoding Network, respectively. 
`ConvBlock' and `DeconvBlock' layers are the same as the `Conv-BN-ReLU' and `Deconv-BN-ReLU' proposed in the FSRNet \cite{FSRNet}.

\begin{table}[!]
\centering
\begin{minipage}[b]{0.414\linewidth}
\centering
\resizebox{\textwidth}{!}{
\begin{tabular}{lll}
\hline
\multicolumn{1}{|c|}{\textbf{Layer Name}} & \multicolumn{1}{c|}{\textbf{Output Size}} & \multicolumn{1}{c|}{\textbf{Structure}} \\ \hline
\multicolumn{1}{|c|}{Input}      & \multicolumn{1}{c|}{128x128}     & \multicolumn{1}{c|}{-}          \\ \hline

\multicolumn{3}{|c|}{\textbf{$f_{enc}$}}                                                            \\ \hline
\multicolumn{1}{|c|}{ConvBlock1} & \multicolumn{1}{c|}{64$\times$64} & \multicolumn{1}{c|}{3$\times$3, 64, stride 2} \\
\multicolumn{1}{|c|}{ResBlock1}  & \multicolumn{1}{c|}{64$\times$64} & \multicolumn{1}{c|}{$\begin{bmatrix} 3\times3, 64\\ 3\times3, 64\\ \end{bmatrix} \times12 $} \\
\multicolumn{1}{|c|}{Conv1}      & \multicolumn{1}{c|}{64$\times$64} & \multicolumn{1}{c|}{3$\times$3, 64, stride 1} \\\hline

\multicolumn{3}{|c|}{\textbf{$f_{dec}$}}                                                            \\ \hline
\multicolumn{1}{|c|}{BatchNorm1}      & \multicolumn{1}{c|}{64$\times$64} & \multicolumn{1}{c|}{-} \\
\multicolumn{1}{|c|}{ConvBlock2}      & \multicolumn{1}{c|}{64$\times$64} & \multicolumn{1}{c|}{3$\times$3, 64, stride 1} \\
\multicolumn{1}{|c|}{DeconvBlock1}    & \multicolumn{1}{c|}{128$\times$128} & \multicolumn{1}{c|}{4$\times$4, 64, stride 2}\\
\multicolumn{1}{|c|}{ResBlock2}       & \multicolumn{1}{c|}{128$\times$128} & \multicolumn{1}{c|}{$\begin{bmatrix} 3\times3, 64\\ 3\times3, 64\\ \end{bmatrix} \times3 $}\\
\multicolumn{1}{|c|}{ConvBlock3}      & \multicolumn{1}{c|}{128$\times$128} & \multicolumn{1}{c|}{3$\times$3, 3, stride 1} \\
\multicolumn{1}{|c|}{Tanh1}           & \multicolumn{1}{c|}{128$\times$128} & \multicolumn{1}{c|}{-} \\\hline

\multicolumn{3}{|c|}{\textbf{$Deconv$}}                                                        \\ \hline
\multicolumn{1}{|c|}{DeconvBlock2}    & \multicolumn{1}{c|}{16$\times$16} & \multicolumn{1}{c|}{4$\times$4, 64, stride 2}\\
\multicolumn{1}{|c|}{ResBlock3}       & \multicolumn{1}{c|}{16$\times$16} & \multicolumn{1}{c|}{$\begin{bmatrix} 3\times3, 64\\ 3\times3, 64\\ \end{bmatrix} $}\\
\multicolumn{1}{|c|}{DeconvBlock3}    & \multicolumn{1}{c|}{32$\times$32} & \multicolumn{1}{c|}{4$\times$4, 64, stride 2}\\
\multicolumn{1}{|c|}{ResBlock4}       & \multicolumn{1}{c|}{32$\times$32} & \multicolumn{1}{c|}{$\begin{bmatrix} 3\times3, 64\\ 3\times3, 64\\ \end{bmatrix} $}\\
\multicolumn{1}{|c|}{DeconvBlock4}    & \multicolumn{1}{c|}{64$\times$64} & \multicolumn{1}{c|}{4$\times$4, 64, stride 2}\\
\multicolumn{1}{|c|}{ResBlock5}       & \multicolumn{1}{c|}{64$\times$64} & \multicolumn{1}{c|}{$\begin{bmatrix} 3\times3, 64\\ 3\times3, 64\\ \end{bmatrix} $}\\\hline
\end{tabular}
}
\caption{The network structure of the SR Encoder-Decoder. It is based on the \textit{Fine SR Encoder} and the \textit{Fine SR Decoder} of the FSRNet \cite{FSRNet}.}
\label{tab:SREncoderDecoder}
\end{minipage}
\hspace{0.5cm}
\begin{minipage}[b]{0.45\linewidth}
\centering
\resizebox{\textwidth}{!}{
\begin{tabular}{lll}
\hline
\multicolumn{1}{|c|}{\textbf{Layer Name}} & \multicolumn{1}{c|}{\textbf{Output Size}} & \multicolumn{1}{c|}{\textbf{Structure}} \\ \hline
\multicolumn{1}{|c|}{Input}      & \multicolumn{1}{c|}{128x128}     & \multicolumn{1}{c|}{-}          \\ \hline

\multicolumn{3}{|c|}{\textbf{$q$}} \\ \hline
\multicolumn{1}{|c|}{ConvBlock4} & \multicolumn{1}{c|}{64$\times$64} & \multicolumn{1}{c|}{3$\times$3, 64, stride 2} \\
\multicolumn{1}{|c|}{ResBlock6}       & \multicolumn{1}{c|}{64$\times$64} & \multicolumn{1}{c|}{$\begin{bmatrix} 3\times3, 64\\ 3\times3, 64\\ \end{bmatrix} $}\\
\multicolumn{1}{|c|}{ConvBlock5} & \multicolumn{1}{c|}{32$\times$32} & \multicolumn{1}{c|}{3$\times$3, 64, stride 2} \\
\multicolumn{1}{|c|}{ResBlock7}       & \multicolumn{1}{c|}{32$\times$32} & \multicolumn{1}{c|}{$\begin{bmatrix} 3\times3, 64\\ 3\times3, 64\\ \end{bmatrix} $}\\
\multicolumn{1}{|c|}{ConvBlock6} & \multicolumn{1}{c|}{16$\times$16} & \multicolumn{1}{c|}{3$\times$3, 64, stride 2} \\
\multicolumn{1}{|c|}{ResBlock8}       & \multicolumn{1}{c|}{16$\times$16} & \multicolumn{1}{c|}{$\begin{bmatrix} 3\times3, 64\\ 3\times3, 64\\ \end{bmatrix} $}\\
\multicolumn{1}{|c|}{ConvBlock7} & \multicolumn{1}{c|}{8$\times$8} & \multicolumn{1}{c|}{3$\times$3, 64, stride 2} \\
\multicolumn{1}{|c|}{ResBlock9}       & \multicolumn{1}{c|}{8$\times$8} & \multicolumn{1}{c|}{$\begin{bmatrix} 3\times3, 64\\ 3\times3, 64\\ \end{bmatrix} $}\\
\multicolumn{1}{|c|}{Conv2}      & \multicolumn{1}{c|}{8$\times$8} & \multicolumn{1}{c|}{3$\times$3, 64, stride 1} \\\hline
\multicolumn{3}{|c|}{\textbf{${\mu}_{pred}, {\sigma}_{pred}$}} \\ \hline
\multicolumn{1}{|c|}{Conv3}      & \multicolumn{1}{c|}{8$\times$8} & \multicolumn{1}{c|}{3$\times$3, 32, stride 1} \\
\multicolumn{1}{|c|}{BatchNorm2} & \multicolumn{1}{c|}{8$\times$8} & \multicolumn{1}{c|}{-} \\
\multicolumn{1}{|c|}{LeakyReLU1} & \multicolumn{1}{c|}{8$\times$8} & \multicolumn{1}{c|}{-} \\
\multicolumn{1}{|c|}{Conv4}      & \multicolumn{1}{c|}{8$\times$8} & \multicolumn{1}{c|}{3$\times$3, 64, stride 1} \\
\multicolumn{1}{|c|}{BatchNorm3} & \multicolumn{1}{c|}{8$\times$8} & \multicolumn{1}{c|}{-} \\
\multicolumn{1}{|c|}{Tanh2}      & \multicolumn{1}{c|}{8$\times$8} & \multicolumn{1}{c|}{-} \\\hline
\end{tabular}
}
\caption{The network structure of the Input Conditional Attribute Predictor. For simplicity of training, it makes log of $\vect{\sigma}_{pred}^2$ instead of $\vect{\sigma}_{pred}$.}
\label{tab:ICAP}
\end{minipage}
\end{table}

\begin{table}[!]
\begin{center}
\resizebox{0.45\textwidth}{!}{\begin{tabular}{lll}
\hline
\multicolumn{1}{|c|}{\textbf{Layer Name}} & \multicolumn{1}{c|}{\textbf{Output Size}} & \multicolumn{1}{c|}{\textbf{Structure}} \\ \hline
\multicolumn{1}{|c|}{Input}      & \multicolumn{1}{c|}{128x128}     & \multicolumn{1}{c|}{-}          \\ \hline

\multicolumn{3}{|c|}{\textbf{$g$}} \\ \hline
\multicolumn{1}{|c|}{ConvBlock8} & \multicolumn{1}{c|}{8$\times$8} & \multicolumn{1}{c|}{[3$\times$3, 64, stride 2] $\times$4} \\
\multicolumn{1}{|c|}{Conv5}      & \multicolumn{1}{c|}{8$\times$8} & \multicolumn{1}{c|}{3$\times$3, 64, stride 1} \\
\multicolumn{1}{|c|}{BatchNorm4} & \multicolumn{1}{c|}{8$\times$8} & \multicolumn{1}{c|}{-} \\\hline

\multicolumn{3}{|c|}{\textbf{${\mu}_{res}, {\sigma}_{res}$}} \\ \hline
\multicolumn{1}{|c|}{Conv6}      & \multicolumn{1}{c|}{8$\times$8} & \multicolumn{1}{c|}{3$\times$3, 32, stride 1} \\
\multicolumn{1}{|c|}{BatchNorm5} & \multicolumn{1}{c|}{8$\times$8} & \multicolumn{1}{c|}{-} \\
\multicolumn{1}{|c|}{LeakyReLU2} & \multicolumn{1}{c|}{8$\times$8} & \multicolumn{1}{c|}{-} \\
\multicolumn{1}{|c|}{Conv7}      & \multicolumn{1}{c|}{8$\times$8} & \multicolumn{1}{c|}{3$\times$3, 64, stride 1} \\
\multicolumn{1}{|c|}{BatchNorm6} & \multicolumn{1}{c|}{8$\times$8} & \multicolumn{1}{c|}{-} \\
\multicolumn{1}{|c|}{Tanh3}      & \multicolumn{1}{c|}{8$\times$8} & \multicolumn{1}{c|}{-} \\\hline

\end{tabular}}
\end{center}
\caption{The network structure of the Residual Encoding Network. For simplicity of training, it makes log of $\vect{\sigma}_{res}^2$ instead of $\vect{\sigma}_{res}$.}
\label{tab:REN}
\end{table}

\end{document}